\documentclass[10pt,twocolumn,letterpaper]{article}

\usepackage{cvpr}              

\usepackage{graphicx}
\usepackage{amsmath}
\usepackage{amssymb}
\usepackage{booktabs}

\usepackage{multirow}
\usepackage{booktabs}
\usepackage{makecell}
\usepackage{xcolor}
\usepackage{colortbl}
\usepackage{wrapfig}
\usepackage{floatrow}
\usepackage{caption}
\usepackage{subcaption}
\usepackage{bbding}
\usepackage{float}
\usepackage{pifont}
\newcommand{\cmark}{\ding{51}}%
\newcommand{\xmark}{\ding{53}}%
\usepackage[misc]{ifsym}
\usepackage{setspace}

\usepackage[pagebackref,breaklinks,colorlinks,bookmarks=false]{hyperref}

\usepackage[capitalize]{cleveref}
\crefname{section}{Sec.}{Secs.}
\Crefname{section}{Section}{Sections}
\Crefname{table}{Table}{Tables}
\crefname{table}{Tab.}{Tabs.}

\def\cvprPaperID{7777} 
\def\confName{CVPR}
\def\confYear{2023}

\begin{document}

\title{Learning Transferable Spatiotemporal Representations \\ from Natural Script Knowledge}

\author{
Ziyun Zeng\textsuperscript{1,2 *}\quad 
Yuying Ge\textsuperscript{3 *}\quad 
Xihui Liu\textsuperscript{3}\quad \\
Bin Chen\textsuperscript{4 \Letter}\quad
Ping Luo\textsuperscript{3}\quad
Shu-Tao Xia\textsuperscript{1}\quad
Yixiao Ge\textsuperscript{2 \Letter} \\
$^1$ Tsinghua University\quad 
$^2$ Applied Research Center (ARC), Tencent PCG\quad \\
$^3$ The University of Hong Kong\quad
$^4$ Harbin Institute of Technology, Shenzhen \\
{\small
\textsuperscript{*} equal contribution\quad
\textsuperscript{\Letter} corresponding authors
} \\
{\tt\small
zengzy21@mails.tsinghua.edu.cn\quad
yuyingge@hku.hk\quad
xihuiliu@eee.hku.hk
} \\
{\tt\small
chenbin2021@hit.edu.cn\quad
pluo@cs.hku.hk\quad
xiast@sz.tsinghua.edu.cn\quad
yixiaoge@tencent.com
} \\
}
\maketitle


\begin{abstract}
Pre-training on large-scale video data has become a common recipe for learning transferable spatiotemporal representations in recent years.
Despite some progress, existing methods are mostly limited to highly curated datasets (\eg, K400) and exhibit unsatisfactory out-of-the-box representations.
We argue that it is due to the fact that they only capture pixel-level knowledge rather than spatiotemporal semantics, which hinders further progress in video understanding.
Inspired by the great success of image-text pre-training (\eg, CLIP), we take the first step to exploit language semantics to boost transferable spatiotemporal representation learning.
We introduce a new pretext task, Turning to Video for Transcript Sorting (TVTS), which sorts shuffled ASR scripts by attending to learned video representations.
We do not rely on descriptive captions and learn purely from video, \ie, leveraging the natural transcribed speech knowledge to provide noisy but useful semantics over time. Our method enforces the vision model to contextualize what is happening over time so that it can re-organize the narrative transcripts, and can seamlessly apply to large-scale uncurated video data in the real world.
%
Our method demonstrates strong out-of-the-box spatiotemporal representations on diverse benchmarks, \eg, +13.6\% gains over VideoMAE on SSV2 via linear probing.
The code is available at \url{https://github.com/TencentARC/TVTS}.
\end{abstract}
\section{Introduction}




The aspiration of representation learning is to encode general-purpose representations that transfer well to diverse downstream tasks, where self-supervised methodologies \cite{moco,simclr} dominate due to their advantage in exploiting large-scale unlabeled data. 
Despite significant progress in learning representations of still images \cite{mae,CLIP}, the real world is dynamic and requires reasoning over time.
In this paper, we focus on \emph{out-of-the-box spatiotemporal representation learning}, a more challenging but practical task towards generic video understanding, which aims to capture hidden representations that can be further used to conduct reasoning on broader tasks, \eg, classification and retrieval.

\begin{figure*}[t]
	\centering
	\includegraphics[width=1.0\linewidth]{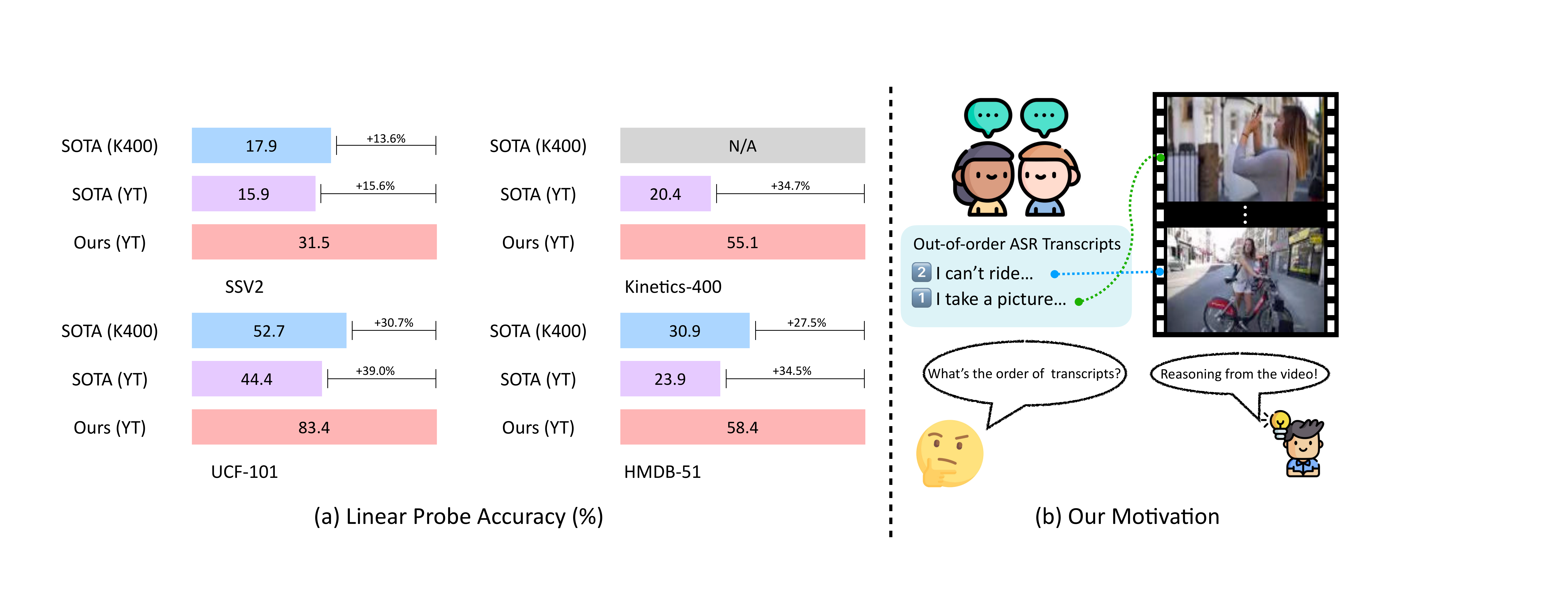}
	\caption{(a) We evaluate the transferability of spatiotemporal representations via linear probing on four video recognition datasets \cite{ssv2,kinetics,ucf,hmdb}, where the state-of-the-art method \cite{VideoMAE} underperforms. It performs even worse when pre-trained with a large-scale uncurated dataset, YT-Temporal \cite{merlot}.
	(b) We encourage complex temporal understanding and advanced spatiotemporal representation learning with a new pretext task of sorting transcripts.}
	\label{fig:intro}
\end{figure*}

There have been various attempts at self-supervised pre-training on video data from discriminative learning objectives~\cite{rspnet,ascnet,long} to generative ones~\cite{VideoMAE,VideoMAE_Kaiming}, where the core is context capturing in spatial and temporal dimensions.
Though promising results are achieved when transferring the pre-trained models to downstream video recognition~\cite{ssv2,ucf,hmdb} via fine-tuning, the learned representations are still far away from out-of-the-box given the poor linearly probing results (see Figure \ref{fig:intro}(a)).
Moreover, existing works mostly develop video models on the highly curated dataset with particular biases, \ie, K400~\cite{kinetics}.
Their applicability in the real world is questioned given the observed performance drops when training on a larger but uncurated dataset, YT-Temporal~\cite{merlot}.
We argue that, to address the above issue, the rich spatiotemporal semantics contained in the video itself should be fully exploited. But current video models generally exploit visual-only perception (\eg, pixels) without explicit semantics.

%
%
Recently, the success of CLIP~\cite{CLIP} has inspired the community to learn semantically aware image representations that are better transferable to downstream tasks and scalable to larger uncurated datasets.
It provides a feasible solution for improving spatiotemporal representation learning but remains two key problems.
(1) The vision-language contrastive constraints in CLIP mainly encourage the understanding of static objects (noun contrast) and simple motions (verb contrast), while how to enable long-range temporal understanding with language supervision needs to be studied.
(2) The quality of language supervision \cite{santurkar2022caption} is critical to the final performance of CLIP, however, it is hard to collect large-scale video data with literal captions that carefully describe the dynamic content over time.
The ideal way for self-supervised learning is to learn useful knowledge purely from the data itself, which is also the philosophy followed by previous video pre-training methods \cite{VideoMAE,VideoMAE_Kaiming}. Fortunately, video data is naturally multi-modal with transcribed speech knowledge in the form of text (ASR), providing time-dependent semantics despite some noise.
To facilitate spatiotemporal understanding in large-scale uncurated data under the supervision of inherent script knowledge,
we introduce
a new pretext task for video pre-training, namely, {\textbf{T}urning to \textbf{V}ideo for \textbf{T}ranscript \textbf{S}orting (TVTS)}.
Intuitively, people sort out the order of events by temporal reasoning.
As illustrated in Figure \ref{fig:intro}(b), given several unordered transcripts, it is difficult to reorganize the narrative by merely understanding the literal semantics. When the corresponding video is provided, it will be much easier to sort the transcripts by contextualizing what is happening over time. 
Whereas in neural networks, the temporal inference is embedded in spatiotemporal representations. 
Thus we believe that if the chronological order of transcripts can be correctly figured out via resorting to the correlated video representations, the video has been well understood.

We realize the pretext task of TVTS by performing joint attention among the encoded video spatiotemporal representations and the extracted ASR transcript representations. 
Specifically, given an input video and its successive transcripts, we randomly shuffle the order of the sentences.
Subsequently, 
we concatenate the encoded script representations and the video representations and perform self-attention to predict the actual orders of the shuffled transcripts by fully understanding the spatiotemporal semantics in the video. 
The order prediction is cast as a $K$-way classification task, where $K$ is the number of transcripts.
The pretext task indirectly regularizes our model to properly capture contextualized spatiotemporal representations to provide enough knowledge for transcript ordering.

%

The usage of language supervision is related to video-text alignment \cite{frozen,mcq} and multimodal representation learning \cite{merlot,violet} methods, however, we are completely different.
(1) Video-text alignment methods focus on retrieval tasks and are devoted to associating the vision patterns with language concepts. They are generally single-frame biased \cite{singularity} and fail to encode strong out-of-the-box temporal representations.
(2) Multimodal representation learning methods aim to learn fused representations across modalities rather than vision-only spatiotemporal representations in our work. Moreover, different from our pretext task that aims to optimize spatiotemporal video representations, \cite{merlot} sorts video frames by taking the features of individual frames as inputs without temporal modeling, \ie, learning video representations only at the image level. 
As \cite{merlot} points out, its ordering pretext task is not critical for downstream tasks (performance even drops) and primarily serves as an interface to query the model about temporal events.

To summarize, our contributions are three-fold.
(\textbf{i}) We exploit the rich semantics from script knowledge which is naturally along with the video, rendering a flexible pre-training method that can easily apply to uncurated video data in the real world. 
(\textbf{ii}) We introduce a novel pretext task for video pre-training, namely, Turning to Video for Transcript Sorting (TVTS). It promotes the capability of the model in learning transferable spatiotemporal video representations. 
(\textbf{iii}) We conduct comprehensive comparisons with advanced methods.
Our pre-trained model exhibits strong out-of-the-box spatiotemporal representations on downstream action recognition tasks, especially the relatively large-scale and the most challenging SSV2~\cite{ssv2}. We also achieve state-of-the-art performances on eight common video datasets in terms of fine-tuning.

\section{Related Work}
\noindent\textbf{Spatiotemporal representation learning.}
Dominant video representation learning works have two categories, \ie, discriminative- and generative-based methods.
\textbf{(i)} The discriminative-based methods aim at mining unique representations within videos. For example, SVT~\cite{SVT} aligns several views from the same video with different spatial and temporal resolution for video-invariant representations. RSPNet~\cite{rspnet}, ASCNet~\cite{ascnet}, and LongShortView~\cite{long} utilize the appearance and temporal consistency of videos as the supervision. They use different augmentations of videos to construct positive and negative pairs to learn correspondences along the spatial and temporal dimensions. 
\textbf{(ii)} The generative-based methods try to reconstruct visual information from corrupted inputs. For example, MAE-based~\cite{ImageMAE} methods~\cite{VideoMAE,VideoMAE_Kaiming} use pixel values of video frames as supervision by masking raw videos with an extremely high ratio and reconstructing them.

Previous works are mainly trained on highly curated datasets, \eg, Kinetics-400, HMDB51, and UCF101, where the temporal motions are not significant~\cite{singularity}. This leads to a ``spatial bias'', thus weakening the transferability to real-world uncurated datasets due to the lack of long-term temporal reasoning. Besides, existing works merely use visual supervision without explicit semantic information. 
Compared to them, our work leverages natural language derived from the video itself, \ie, the ASR transcripts, as the supervision. Benefiting from the rich spatiotemporal information, our learned video representations have stronger transferability to downstream tasks.

\noindent\textbf{Video-text pre-training.}
Existing video-text pre-training work can be divided into two categories. The first category aims to learn video-text alignment for retrieval. For example, Frozen~\cite{frozen}, MCQ~\cite{mcq}, and MILES~\cite{miles} generally adopt two separate encoders to extract video and text representations, then align them with contrastive loss. However, they only align videos with a global video caption, thus neglecting the fine-grained temporal information. Furthermore, they rely on clean captions, which are difficult to scale up, and it is actually hard to collect large-scale video data with captions describing the dynamic content over time.
The second category works on joint representation learning across modalities mainly for VQA. For example, MERLOT~\cite{merlot} adopts a joint encoder to match the captions with the corresponding video frames and put scrambled video frames into the correct order. It aims to match different modalities in the temporal dimension to achieve multi-modality fusion in a joint encoder, rather than learn better spatiotemporal representations.


\noindent\textbf{Image representation learning by language supervision.}
Recently, there have been a bunch of successful tries in utilizing language supervision to enhance image representation learning. 
For example, CLIP~\cite{CLIP} utilized 400M image-text pairs collected from the Internet and adopt the contrastive loss to align the image and its corresponding text. The superior performance on downstream image classification tasks revealed that learning directly from the raw text about images is a promising alternative that leverages a much broader source of supervision. ALIGN~\cite{ALIGN}, uses a larger but noisier uncurated dataset and shows similar results to CLIP. 
Nevertheless, these methods only utilize language supervision to improve spatial learning, without exploring temporal learning, which hinders them from properly learning out-of-the-box video representations.

\section{Method}


\begin{figure*}[t]
	\centering
	\includegraphics[width=1.0\linewidth]{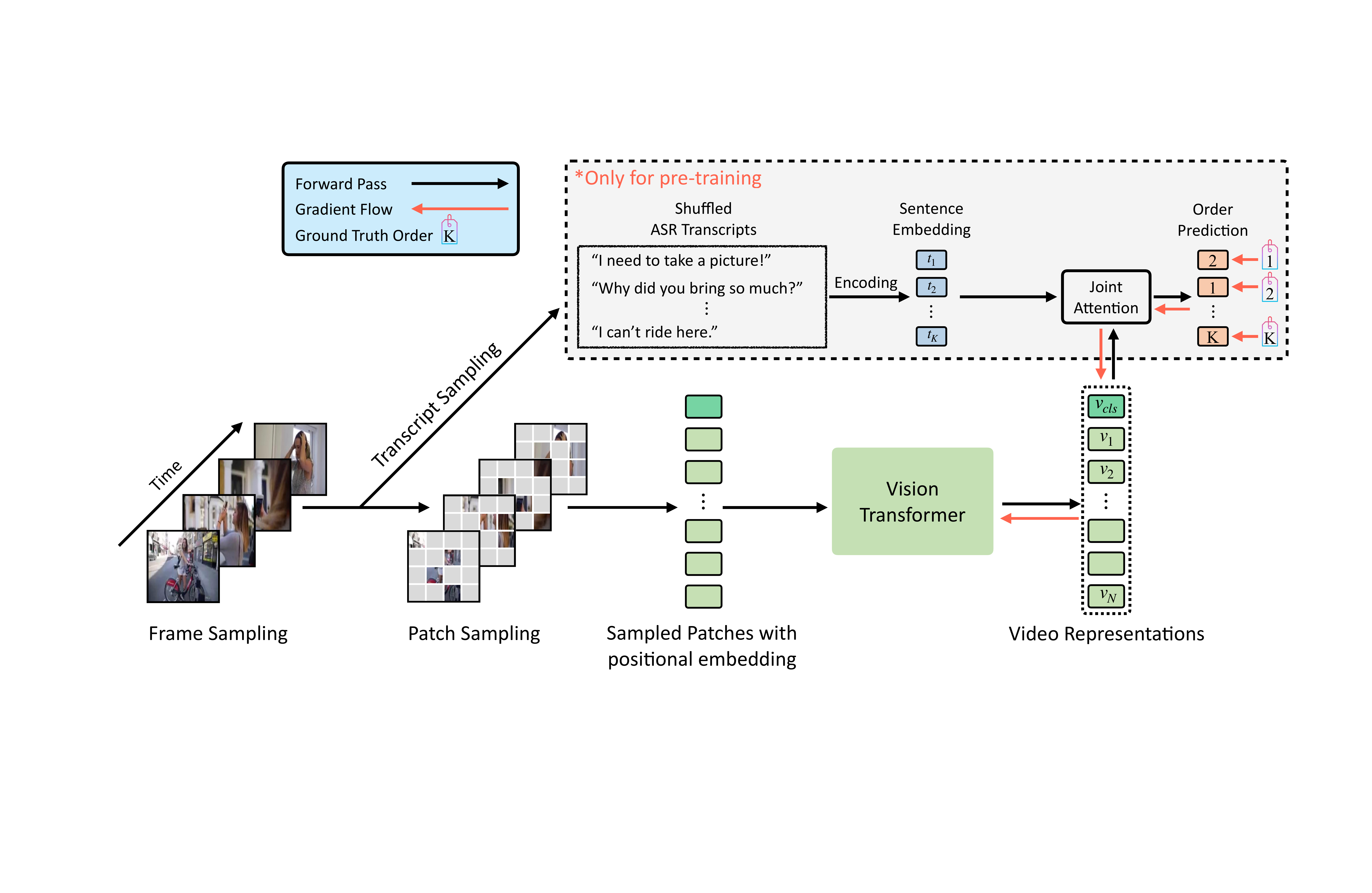}
	\caption{Our pre-training pipeline. We first sample $K$ consecutive ASR transcripts, and a video clip consisting of $M$ frames within the span of the transcripts. We randomly sample frame patches as the input of a vision transformer for the video representations. We then shuffle the transcripts and extract the representation of each transcript. We perform joint attention among the transcript and the video representations to predict the actual order of each transcript, which is optimized with a cross-entropy objective.}
	\label{fig:method}
\end{figure*}

In this work, we introduce a novel pretext task, \textbf{Turning to Video for Transcript Sorting (TVTS)} to learn the transferable spatiotemporal video representation by leveraging the rich semantics from script knowledge. In this section, we first introduce the pretext TVTS in Sec.~\ref{TVTS} and our pre-training objectives in Sec.~\ref{objective}. We then describe the model architecture in Sec.~\ref{model}. 

\subsection{Turning to Video for Transcript Sorting}\label{TVTS}
As shown in Fig.~\ref{fig:method}, we perform the pretext task of TVTS to learn transferable spatiotemporal representations of videos. Given the observation that it will be much easier to sort the ASR transcripts by contextualizing what is happening over time in the video, we first randomly shuffle several consecutive ASR transcripts and extract their representations. We then perform joint attention among the transcript representations and video representations to sort the transcripts in the correct order via capturing contextualized spatiotemporal representations of videos. 


{\flushleft \bf Sample and Shuffle.} 
Given a video $V$ and its corresponding ASR transcripts with word-level timestamps $\{(w_i,s_i)\}_{i=1}^{N_{\text{asr}}}$, where $N_{\text{asr}}$ denotes the word number, $w_i$ and $s_i$ denote the \emph{i}-th word and its timestamp respectively, we randomly choose a starting time $s_{\text{begin}}$ and sample $K$ consecutive transcripts, each with a duration of $l$ (in seconds), and an interval of 1s between adjacent transcripts,  
\begin{equation}
    \begin{split}
    &S_k = s_{\text{begin}}+(k-1)*(l+1),\quad E_k = S_k + l \\
    &T_k = \{w_i|S_k \le s_i \le E_k\},\quad k\in\{1,\cdots, K\}
    \end{split}
\label{eqn:sample}
\end{equation}
where $S_k$ and $E_k$ denote the beginning and ending time of the \emph{k}-th transcript. We consecutively sample $K$ transcripts with an interval of 1s and collect all words within $[S_k,E_k]$ for the \emph{k}-th transcript.  Finally, we randomly shuffle the transcripts as $\{T_{o_i}\}_{i=1}^K$, which means that the \emph{i}-th transcript in this shuffled sequence is actually the $o_i$-th transcript in the original ordered sequence.

As for the video, we sample a clip between the beginning and ending time of all $K$ transcripts, \ie, $[S_1,E_K]$, which contains $M$ frames as $\{F_i\}_{i=1}^{M}$. Specifically, we follow TSN~\cite{tsn} to divide $[S_1, E_K]$ into $M$ segments with equal length and randomly sample 1 frame from each segment. After that, we get a video clip with $M$ frames and $K$ shuffled transcripts along the span of the video clip.



{\flushleft \bf Sorting Transcripts.} 
Given the shuffled transcripts $\{T_{o_i}\}_{i=1}^K$ and the corresponding video clip $\{F_i\}_{i=1}^{M}$, we first feed the transcripts in parallel to encode \emph{unordered} text representations $\{t_{o_i}\}_{i=1}^K$. We then randomly sample the frame patches by masking a large proportion of the video clip among the spatial and temporal dimension as the input of a vision transformer to encode video representations $\{v_j\}_{j=0}^N$, where $N$ denotes the number of the unmasked video patches, and $v_0$ is the representation of the [CLS] token. It is worth noting that \textit{we do not add the extra [MASK] token, and we have no explicit reconstruction target}, which is different from previous works \cite{VideoMAE,VideoMAE_Kaiming}. We sample video frame patches as a means of data augmentation since it provides corrupted knowledge for our model to perform the pretext task of TVTS. Such a strategy also reduces the computational cost during pre-training
as the attention is calculated on fewer patches.

We then concatenate the text representations of the shuffled transcripts $\{t_{o_i}\}_{i=1}^K$ and the video representations of the sampled video clip $\{v_j\}_{j=0}^N$, and perform multi-head self-attention among them. Our model attempts to sort the transcripts in the correct order by attending to the text features of all transcripts and the visual features of the unmasked video clip. We model the prediction of the transcript orders as a $K$-way classification task for each transcript. The first $K$ output representations after the joint attention are further fed into a linear classifier to predict the order $p\in\mathbb{R}^K$, where $p_j$ denotes the probability that the transcript is the \emph{j}-th transcript in the original ordered sequence. For the transcript $T_{o_i}$, the ground truth classification label should be $o_i$.

The pretext task of TVTS regularizes the model to contextualize what is happening over time, so that it can provide enough knowledge for our model to figure out the chronological order of the shuffled transcripts. It improves the capability of the model to learn spatiotemporal representations that can be transferred to downstream tasks. We compare our method with other works that also adopt an ordering-based pretext task for pre-training in Sec.~\ref{sec:comparison}.

\subsection{Pre-training Objectives}\label{objective}
Besides the pretext task of TVTS, we use a global video-transcript contrastive objective. It aligns the features of the video clip and the averaged features of $K$ transcripts so that the video and transcript representations are in the same feature space for performing the joint attention to predict transcript orders. We combine two objectives to optimize the entire model in an end-to-end manner.

The first one is a cross-entropy objective $\mathcal{L}_{\text{sort}}$, which supervises our model to predict the correct order of the transcripts, and is formulated as below,
\begin{equation}\small
    \begin{split}
    &\mathcal{L}_{\text{sort}}=-\frac{1}{K}\sum_{i=1}^K
    \log\text{softmax}(\hat{p}^{i}) \\
    s.t.\quad
    &\text{softmax}(\hat{p}^i)=\frac{\exp(p_{o_i}^i)}{\sum_{j=1}^K\exp(p_{j}^i)},
    \end{split}
\end{equation}
where $p^i_{j}$ denotes the probability that the \emph{i}-th transcript in the shuffled sequence is the \emph{j}-th transcript in the original ordered sequence and $o_i$ is the ground truth order in the original ordered sequence.

%
The second one is the global video-transcript contrastive objective $\mathcal{L}_{\text{base}}$, formulated as a bidirectional InfoNCE~\cite{InfoNCE},
\begin{equation}\small
    \begin{split}
    &\mathcal{L}_{\text{base}} = \text{NCE}(\hat{t},\hat{v}) + \text{NCE}(\hat{v},\hat{t}) \\ 
    s.t.\quad
    &\text{NCE}(q,k) = -\log\frac{\exp(q^\top k_+/\tau)}{\sum_{i=1}^B \exp(q^\top k_i/\tau)},
    \end{split}
\end{equation}
where $\hat{t}$ and $\hat{v}$ denote the global text and video representation. We average the [CLS] token representation of all $K$ transcripts as $\hat{t}$, \ie, $\hat{t} \leftarrow \frac{1}{K} \sum_{i=1}^K t_i$, and use the [CLS] token representation of the video clip as $\hat{v}$, \ie, $\hat{v} \leftarrow v_0$.

Our overall pre-training objective combines the two objectives, \ie, $\mathcal{L}  =  \mathcal{L}_{\text{base}} + \lambda \mathcal{L}_{\text{sort}}$,
where $\lambda$ is a hyper-parameter to balance the two losses. 
In our implementation, we set $\lambda=2$ to roughly scale the gradient magnitudes of $\mathcal{L}_{\text{base}}$ and $\mathcal{L}_{\text{sort}}$ to be the same for efficient training.

\subsection{Model Architecture}\label{model}
The vision transformer takes a video clip as input, which consists of $M$ frames of resolution $H\times W$, and outputs video representations. We follow~\cite{VideoMAE} to adopt cube embeddings, where each token corresponds to a cube of size $2\times 16\times 16$. This yields $\frac{M}{2}\times\frac{H}{16}\times \frac{W}{16}$ 3D tokens. Then we add divided space-time embedding to the token sequence, where tokens within the same frame obtain the same temporal embedding, and tokens within the same spatial location of different frames obtain the same spatial embedding. In this way, the vision transformer learns the positional information of the cubes. Next, we follow BERT~\cite{bert} to add a learnable [CLS] token at the beginning of the token sequence for global video representations.
Then we mask a portion of video tokens without [MASK] token replacement, as stated in Sec.~\ref{TVTS}.
We adopt a standard ViT~\cite{ViT} architecture to encode video representations. The unmasked $N$ video tokens as well as the [CLS] token are fed into the vision transformer, and joint space-time attention~\cite{vivit} is performed among the whole unmasked token sequence.

We use a DistilBERT~\cite{distilbert} to extract the representations of ASR transcripts. We adopt two stacked bidirectional transformer blocks to predict the order of each transcript by performing joint attention among the transcript and the video representations. Within each block, multi-head self-attention is performed among all the video and text tokens, \ie, all transcript-video tokens interact with each other.

\section{Experiments}

\begin{table*}[t]
\centering
\caption{Comparison with methods that use ordering-based pretext tasks for pre-training. We report top-1 accuracy under the linear probe classification protocol on UCF-101 and HMDB-51. The model pre-trained only with $\mathcal{L}_{\text{base}}$ serves as the baseline. All models are pre-trained on the YT-Temporal dataset for a fair comparison.}
\scalebox{1.0}{
\begin{tabular}{cccccc}
\toprule
Target $\rightarrow$ & None     & Transcript & \multicolumn{3}{c}{Video}            \\
\midrule
Dataset $\downarrow$     & Baseline & Ours   & VCOP~\cite{VCOP}  & MERLOT~\cite{merlot} & MERLOT-like \\
\midrule
UCF-101     & 81.2 ($\downarrow$2.2)     & \textbf{83.4}            & 79.1 ($\downarrow$4.3)          & 74.9 ($\downarrow$8.5)   & 80.1 ($\downarrow$3.3) \\
HMDB-51     & 56.5 ($\downarrow$1.9)     & \textbf{58.4}           & 54.2 ($\downarrow$4.2)            & 49.6 ($\downarrow$8.8)  & 55.4 ($\downarrow$3.0) \\
\bottomrule
\end{tabular}}
\label{tab:sort_pretext_tasks}
\end{table*}

\subsection{Pre-training Datasets}
We pre-train our model on the large-scale \textbf{YT-Temporal} dataset~\cite{merlot} containing 6M YouTube videos with ASR transcripts and word-level timestamps. 

%

\subsection{Downstream Tasks}
{\flushleft \bf Action Recognition.} 
We evaluate our pre-trained model on four common video datasets: 
(a) \textbf{Something-Something V2} (SSV2)~\cite{ssv2}, (b) \textbf{Kinetics-400} (K400)~\cite{kinetics}, (c) \textbf{UCF-101}~\cite{ucf}, (d) \textbf{HMDB-51}~\cite{hmdb}. 
Our evaluation is two-fold:
(i) We conduct \emph{zero-shot} \textbf{video-to-video retrieval} and \emph{linear probe} \textbf{classification} on SSV2 to evaluate the transferability of the learned video representation.
The former aims to retrieve videos of the same category as a query video, and the latter freezes the visual encoder and only optimizes a linear classifier.
(ii) We \emph{fully fine-tune} our pre-trained model on the training set of the four datasets to evaluate the action recognition capability. See Appendix for details.


{\flushleft \bf Text-to-Video Retrieval.} 
Beyond action recognition, 
we further evaluate retrieval performance on four benchmarks to see if the improved semantic-aware video representation can benefit retrieval tasks: 
(a) \textbf{MSR-VTT}~\cite{msrvtt} (b) \textbf{DiDeMo}~\cite{didemo} (c) \textbf{MSVD}~\cite{msvd} (d) \textbf{LSMDC}~\cite{lsmdc}.
We adopt Recall@K (R@K) and Median Rank (MedR) as the evaluation metric. See Appendix for details.

\begin{table}[t]
\centering
\resizebox{1.\columnwidth}{!}{
\begin{tabular}{cccccc}
\toprule
Name & $\mathcal{L}_{\text{base}}$ & $\mathcal{L}_{\text{sort}}$  & sg &SSV2   & Kinetics-400   \\
\midrule
M$_\text{scratch}$ & \xmark        & \xmark & -      & 64.5 ($\downarrow$4.0)   & 75.4 ($\downarrow$3.4)   \\
M$_\text{base}$ & \cmark        & \xmark  & -     & 67.0 ($\downarrow$1.5)     & 77.8 ($\downarrow$1.0)   \\
M$_\text{sort\textbackslash sg}$ & \xmark        & \cmark  & \xmark   & failed & failed \\
M$_\text{sort}$ & \xmark        & \cmark  & \cmark   & failed & failed \\
M$_\text{ours\textbackslash sg}$ &\cmark        & \cmark  & \xmark     & 66.2 ($\downarrow$2.3)   & 76.5 ($\downarrow$2.3)   \\
M$_\text{ours}$ & \cmark        & \cmark  & \cmark     & \textbf{68.5}   & \textbf{78.8}   \\
\bottomrule
\end{tabular}}
\caption{The top-1 accuracy under the fine-tuning protocol on SSV2 and Kinetics-400, \wrt different pre-training objectives, where $\mathcal{L}_{\text{sort}}$ trains the pretext task of TVTS. sg denotes stopping gradients of $\mathcal{L}_{\text{sort}}$ towards encoding transcript representations.}
\label{tab:objective}
\end{table}

\begin{table}[t]
\setlength\tabcolsep{4pt}
\centering
\caption{The top-1 accuracy under the fine-tuning protocol on SSV2 and Kinetics-400, \wrt different ways to model the sorting of transcripts. ``Pairwise'' predicts the relative order for all transcript pairs, and ``Factorial'' performs $K!$-classification for all possible orders. Our method uses ``$K$-way'' classification to predict the order of each transcript. The model pre-trained with $\mathcal{L}_{\text{base}}$ only serves as the baseline.}
\resizebox{1.\columnwidth}{!}{
\begin{tabular}{ccccc}
\toprule
\multirow{2}{*}{Dataset}&None & \multicolumn{3}{c}{Sort Modeling}            \\
\cmidrule(l){2-5}
                        &Baseline & Pairwise & Factorial &$K$-way \\
\midrule
SSV2    &67.0 ($\downarrow$1.5)& 67.4 ($\downarrow$1.1)     & 67.2 ($\downarrow$1.3)           & \textbf{68.5} \\
K400    &77.8 ($\downarrow$1.0) & 78.1 ($\downarrow$0.7)     & 78.0 ($\downarrow$0.8)                & \textbf{78.8} \\
\bottomrule
\end{tabular}}
\label{tab:sort_proxy}
\end{table}

\subsection{Implementation Details}\label{sec:iml_details}
We follow recent works~\cite{frozen,mcq} to adopt the pre-trained DistilBERT~\cite{distilbert} to extract transcript representations. The vision transformer is a vanilla ViT-Base~\cite{ViT} with patch size $P$=16 and hidden state dimension $D$=768, and is initialized with ImageMAE-Base~\cite{mae}. We set the temperature parameter $\tau$ to be 0.05. We pre-train our model on the YT-Temporal dataset sampling 16 frames for 20 epochs. We randomly mask 75\% tokens within each frame. The input frame is first resized to 256 $\times$ 256, then we apply RandomCrop during training and CenterCrop during inference. The final input resolution of each frame is 224 $\times$ 224. For downstream tasks, we sample 16 frames for action recognition following ~\cite{VideoMAE} and 4 frames for text-to-video retrieval following ~\cite{frozen}. More hyper-parameters are listed in Appendix.

\subsection{Ablation Study}\label{ablation}

\begin{table*}[t]\small
\centering
\caption{Transferability evaluation on SSV2. We report Recall@K for zero-shot video-to-video retrieval and top-1 accuracy for linear probe classification, where video-to-video retrieval aims to retrieve videos of the same category as a query video. $\dagger$ denotes pre-training on YT-Temporal for a fair comparison, and $\ddagger$ denotes the use of official pre-trained weights for evaluation.}
\scalebox{1.0}{
\begin{tabular}{ccccccc}
\toprule
\multirow{2}{*}{Method} & \multirow{2}{*}{Venue} & \multirow{2}{*}{Pre-train Dataset} & \multicolumn{3}{c}{Zero-shot   Video-to-video Retrieval} & \multirow{2}{*}{Linear Probe} \\
\cmidrule(l){4-6}
                        &             &                      & R@1           & R@5            & R@10           &                               \\
\midrule
\multicolumn{7}{l}{\textcolor{gray}{\textit{Spatiotemporal representation learning method(s)}}} \\
CVRL~\cite{cvrl}       &    CVPR'21       & Kinetics-400 & -             & -              & -              & 11.4 ($\downarrow$20.1)                   \\
MViT~\cite{mvit} & ICCV'21 & Kinetics-400 & - & - & - & 19.4 ($\downarrow$12.1) \\
SCVRL~\cite{scvrl} & CVPRW'22 & Kinetics-400 & - & - & - & 13.8 ($\downarrow$17.7) \\
SVT~\cite{SVT}        &    CVPR'22         & Kinetics-400 & 11.3 ($\downarrow$3.4)             & 30.7 ($\downarrow$7.7)              & 41.1 ($\downarrow$9.4)              & 18.3 ($\downarrow$13.2)                   \\
SVT$^\dagger$~\cite{SVT}        &    CVPR'22         & YT-Temporal & 9.9 ($\downarrow$4.8)             & 26.2 ($\downarrow$12.2)              & 36.3 ($\downarrow$14.2)              & 18.0 ($\downarrow$13.5)                   \\
VideoMAE~\cite{VideoMAE}     & NeurIPS'22      & Kinetics-400 & 7.9 ($\downarrow$6.8)     & 18.6 ($\downarrow$19.8)    & 26.5 ($\downarrow$24.0)    & 17.9 ($\downarrow$13.6)                   \\
VideoMAE$^\dagger$~\cite{VideoMAE}      & NeurIPS'22        & YT-Temporal & 7.2 ($\downarrow$7.5)     & 17.6 ($\downarrow$20.8)    & 25.6 ($\downarrow$24.9)    & 15.9 ($\downarrow$15.6)                   \\
\midrule
\multicolumn{7}{l}{\textcolor{gray}{\textit{Video-text alignment method(s)}}} \\
Frozen$^\ddagger$~\cite{frozen}        &     ICCV'21       & CC3M, WebVid-2M & 10.4 ($\downarrow$4.3)    & 28.5 ($\downarrow$9.9)     & 38.7 ($\downarrow$11.8)    & 17.5 ($\downarrow$14.0)                   \\
MCQ$^\ddagger$~\cite{mcq}     & CVPR'22        & CC3M, WebVid-2M       & 10.4 ($\downarrow$4.3)    & 28.6 ($\downarrow$9.8)     & 38.5 ($\downarrow$12.0)    & 18.0 ($\downarrow$13.5)                   \\
MILES$^\ddagger$~\cite{miles}     & ECCV'22              & CC3M, WebVid-2M & 10.3 ($\downarrow$4.4)    & 28.4 ($\downarrow$10.0)    & 38.4 ($\downarrow$12.1)    & 18.6 ($\downarrow$12.9)                   \\
\midrule
\multicolumn{7}{l}{\textcolor{gray}{\textit{Image representation learning method(s)}}} \\
CLIP~\cite{CLIP}       & ICML'21         & WIT & 10.5 ($\downarrow$4.2)    & 28.8 ($\downarrow$9.6)     & 38.8 ($\downarrow$11.7)    & 16.4 ($\downarrow$15.1)                   \\
\midrule
Ours      & CVPR'23          & YT-Temporal & \textbf{14.7}          & \textbf{38.4}           & \textbf{50.5}           & \textbf{31.5}  \\
\bottomrule
\end{tabular}}
\label{tab: transfer}
\end{table*}

{\flushleft \bf Pre-training Objectives.}
To demonstrate the effectiveness of our pretext task TVTS, we pre-train models with different objectives on the YT-Temporal dataset and evaluate them on SSV2 and K400.
The results are listed in Table~\ref{tab:objective}, in which M$_\text{scratch}$ denotes that we directly fine-tune the ImageMAE~\cite{mae} initialized model.
We have the following observations:
(\textbf{i}) M$_\text{base}$ outperforms M$_\text{scratch}$, which indicates that the natural language can be a promising supervision for video representation learning. 
(\textbf{ii}) Compared to M$_\text{base}$, M$_\text{ours}$ further boosts performance by 1.5\%, which demonstrates that TVTS can effectively regularize our model to learn transferable spatiotemporal representations.
(\textbf{iii}) M$_\text{ours\textbackslash sg}$ drops performance, because when the gradients of $\mathcal{L}_{\text{sort}}$ flow towards encoding transcript representations, the model optimizes the transcript representations to ease the ordering task rather than enhance spatiotemporal representations to provide enough knowledge for transcript sorting.
(\textbf{iv}) Both M$_\text{sort\textbackslash sg}$ and M$_\text{sort}$ failed because the parametric module for sorting hardly converges when feeding the misaligned video and transcript representations from distinct latent spaces.


{\flushleft \bf Sort Modeling.}
We explore different ways to model the order prediction of the shuffled transcripts. Besides using $K$-way classification of each transcript, we also tried \emph{Pairwise}, which sorts the transcripts by predicting the relative orders of the ${K(K-1)}/{2}$ transcript pairs, and \emph{Factorial}, which predicts an overall ordering distribution by performing a $K!$-way classification ($K!$ possible orders given $K$ transcripts). As listed in Table~\ref{tab:sort_proxy}, both \emph{Pairwise} and \emph{Factorial} drop performance, because the former ignores the overall relationship among the transcripts while the latter imposes the same penalty on the results when different number of transcripts are sorted incorrectly. But they still outperform the baseline, indicating that sorting transcripts does benefit spatiotemporal representation learning. Our separate $K$-way classification modeling achieves the best performance.

\subsection{Comparison with Ordering-based Pre-training}
\label{sec:comparison}
MERLOT~\cite{merlot} also adopts an ordering-based pretext task, but has a totally different approach and purpose. MERLOT reorders scrambled video frames given the representations of every single frame and the ordered ASR transcripts with a joint encoder, and reserves the joint encoder for downstream multimodal tasks such as VQA. Specifically, MERLOT predicts the relative order of two video frames by binary classification. MERLOT aims to promote the joint encoder in learning multi-modal representations rather than spatiotemporal representations of videos. Since the visual encoder takes a single frame as input, it only achieves semantic understanding at the single-frame level without temporal reasoning among frames. As MERLOT points out, the ordering pretext task is not critical for downstream tasks (performance even drops) and primarily serves as an interface to query the model about temporal events.  We further tailor MERLOT for our architecture as MERLOT-like, which sorts $K$ shuffled video frames by $K$-way classification with the knowledge of the ordered transcripts. Similar to MERLOT, VCOP~\cite{VCOP} also predicts the order of the shuffled video clips, but only pre-trains on videos without language semantics.

As listed in Table~\ref{tab:sort_pretext_tasks}, all models that sort shuffled video clips/frames as the pretext task (\ie, VCOP, MERLOT) perform even worse than the baseline counterpart without sorting.
It indicates that \textbf{sorting shuffled videos in pre-training is infeasible and counterintuitive for improving spatiotemporal representations}, because it does not regularize the video encoder for spatial and temporal reasoning given only a single frame or a short video segment as input. They mainly regularize an extra module beyond the video encoder to figure out the chronological order of the videos. 
By contrast, our pretext task regularizes the model to sort transcripts via reasoning among the video representations, which enforces the model to capture contextualized spatiotemporal representations so that it can provide enough knowledge for transcript ordering.


\subsection{Main Results}

\begin{table}[t]\small
\setlength\tabcolsep{2pt}
\centering
\caption{The top-1 accuracy under the fine-tuning protocol on SSV2 and Kinetics-400. OmniVL adopts a mixture of eight datasets. $\dagger$ denotes pre-training on YT-Temporal, and $\ddagger$ denotes the use of official pre-trained weights for evaluation.}
\resizebox{.95\columnwidth}{!}{
\begin{tabular}{ccccccc}
\toprule
Method                   & Backbone & Pre-train Dataset                        & SSV2        & K400 \\
\midrule
TSM~\cite{tsm}                              & R50 $\times$ 2  & ImageNet-1K        &                  66.0  & -            \\
Vi$^2$CLR~\cite{vi2clr}         & S3D      & Kinetics-400                                     & -           & 71.2   \\
CORP~\cite{corp}                             & R3D-50   & Kinetics-400                                    & 48.8 & -            \\
MoCo v3~\cite{mocov3}                          & ViT-B    & Kinetics-400                                    & 62.4  & -            \\
TANet~\cite{tanet}                            & R50 $\times$ 2  & ImageNet-1K                           & 66.0  & -            \\
MViT~\cite{mvit}                             & ViT-B    & Kinetcis-400                           & 64.7  & 78.4   \\
TimeSformer~\cite{timesformer}                      & ViT-B    & ImageNet-21K                         & 59.5  & 78.3   \\
RSANet~\cite{rsanet}                           & R50      & ImageNet-1K                         & 66.0  & -            \\
SVT~\cite{SVT}                              & ViT-B    & Kinetics-400                                   & 59.2  & 78.1   \\
VideoMAE$^\dagger$~\cite{VideoMAE}                         & ViT-B    & YT-Temporal                   & 67.9  & 78.2   \\
\midrule
Frozen$^\ddagger$~\cite{frozen}                       & ViT-B    & CC3M, WebVid2M                & 55.1 & 76.9   \\
MCQ$^\ddagger$~\cite{mcq}                          & ViT-B    & CC3M, WebVid2M                & 51.5 & 77.8   \\
MILES$^\ddagger$~\cite{miles}                        & ViT-B    & CC3M, WebVid2M                & 54.1 & 77.4   \\
OmniVL~\cite{omnivl}                       & ViT-B    & *Enormous   Datasets                     & 61.6  & 79.1   \\
\midrule
CLIP~\cite{CLIP}                         & ViT-B    & WIT          & 36.3 & 75.2       \\
\midrule
Ours                         & ViT-B    & YT-Temporal                    & 68.5 & 78.8   \\
Ours                         & ViT-B    & \makecell[c]{YT-Temporal \\ CC3M,   WebVid2M} & \textbf{69.1} & \textbf{79.8} \\
\bottomrule
\end{tabular}}
\label{tab: ssv2_k400}
\end{table}

\begin{table*}[t]\small
\centering
\caption{The R@1 and MedR under the fine-tuning protocol on MSR-VTT, DiDeMo, MSVD, and LSMDC for text-to-video retrieval.}
\resizebox{.95\linewidth}{!}{
\begin{tabular}{ccc|ccc|ccc|ccc}
\toprule
\multicolumn{3}{c}{MSR-VTT} & \multicolumn{3}{c}{DiDeMo} & \multicolumn{3}{c}{MSVD} & \multicolumn{3}{c}{LSMDC} \\
\midrule
Method      & R@1   & MedR  & Method     & R@1   & MedR  & Method     & R@1  & MedR & Method    & R@1   & MedR  \\
\midrule
MMT~\cite{mmt}         & 26.6  & 4.0   & CE~\cite{ce}         & 16.1  & 8.3   & NoiseEst~\cite{noiseest}   & 20.3 & 6.0  & NoiseEst~\cite{noiseest}  & 6.4   & 39.0  \\
SupportSet~\cite{supportset}  & 30.1  & 3.0   & ClipBert~\cite{clipbert}   & 20.4  & 6.0   & SupportSet~\cite{supportset} & 28.4 & 4.0  & MMT~\cite{mmt}       & 12.9  & 19.3  \\
Frozen~\cite{frozen}      & 31.0  & 3.0   & Frozen~\cite{frozen}     & 31.0  & 3.0   & Frozen~\cite{frozen}     & 45.6 & 2.0  & Frozen~\cite{frozen}    & 15.0  & 20.0  \\
Ours        & \textbf{34.6}  & \textbf{3.0}   & Ours       & \textbf{32.4}  & \textbf{3.0}   & Ours       & \textbf{45.9} & \textbf{2.0}  & Ours      & \textbf{17.2}  & \textbf{17.0}  \\
\bottomrule
\end{tabular}
}
\label{tab: retri}
\end{table*}

\begin{table}[t]\small
\setlength\tabcolsep{4pt}
\centering
\caption{The top-1 accuracy under the fine-tuning protocol on UCF-101 and HMDB-51. $\dagger$ denotes pre-training on YT-Temporal, and $\ddagger$ denotes the use of official pre-trained weights for evaluation.}
\scalebox{1.0}{
\begin{tabular}{cccc}
\toprule
Method    & Backbone               & UCF-101 & HMDB-51 \\
\midrule
BE~\cite{be}      & I3D                 & 87.1    & 56.2    \\
CMD~\cite{cmd}    & R(2+1)D-26                  & 85.7    & 54.0    \\
Vi$^2$CLR~\cite{vi2clr} &S3D & 89.1    & 55.7    \\
ASCNet~\cite{ascnet}    &S3D-G               & 90.8    & 60.5    \\
TEC~\cite{tec}       & S3D-G               & 88.2    & 63.5    \\
LSFD~\cite{long}     & C3D                & 79.8    & 52.1    \\
MCN~\cite{mcn}       & R3D               & 89.7    & 59.3    \\
TCLR~\cite{tclr}       & R(2+1)D-18              & 84.3    & 54.2    \\
SVT~\cite{SVT}        & ViT-B              & 93.7    & 67.2    \\
VideoMAE$^\dagger$~\cite{VideoMAE}   & ViT-B              & 94.2    & 68.4    \\
\midrule
Frozen$^\ddagger$~\cite{frozen}   & ViT-B                & 91.4    & 65.6    \\
MCQ$^\ddagger$~\cite{mcq}      & ViT-B                & 92.9    & 65.1    \\
MILES$^\ddagger$~\cite{miles}   & ViT-B                 & 92.1    & 66.8    \\
\midrule
Ours        & ViT-B             & \textbf{95.1}    & \textbf{70.5}    \\
\bottomrule
\end{tabular}}
\label{tab: ucf_hmdb}
\end{table}

\subsubsection{Action Recognition}
{\flushleft \bf Out-of-the-box Representations.}
To explore the transferability of the learned video representation, we evaluate zero-shot video-to-video retrieval and linear probe classification. 
We compare our proposed method with seven state-of-the-art methods, including: 
(a) Five video representation learning methods, \ie, CVRL~\cite{cvrl}, MViT~\cite{mvit}, SCVRL~\cite{scvrl}, SVT~\cite{SVT}, and VideoMAE~\cite{VideoMAE}.
(b) Three video-text alignment methods, \ie, Frozen~\cite{frozen}, MCQ~\cite{mcq}, and MILES~\cite{miles}.
(c) One image representation learning method with natural language supervision, \ie, CLIP~\cite{CLIP}. We average frame features as its video representation.

The results are listed in Table~\ref{tab: transfer} and we have the following observations: 
\textbf{(i)} Our method surpasses all baselines by a large margin under all evaluation metrics, which indicates that our learned video representation has stronger transferability that can be used for out-of-domain video recognition.
\textbf{(ii)} Previous video representation learning works yield weak transferability with only visual supervision. It implies that merely exploiting visual-only perception without explicit semantics hinders spatiotemporal understanding. Furthermore, we observe a significant performance drop on VideoMAE when it pre-trains the model on the large-scale uncurated dataset, \ie, YT-Temporal. By contrast, our pre-trained model achieves promising results, which indicates that TVTS can successfully apply to real-world uncurated video data by exploiting rich semantics from script knowledge.
\textbf{(iii)} Our method also outperforms video-text alignment works by a large margin. We infer that these works only focus on alignment between global video and caption representation without exploring fine-grained temporal information. On the contrary, our proposed TVTS regularizes the model to learn transferable spatiotemporal video representations.
\textbf{(iv)} Benefiting from large-scale language supervision, image-based CLIP achieves competitive performance.
But it is still worse than our model because we fully exploit the rich semantics from script knowledge.

{\flushleft \bf Fine-tuning Transferability.}
We evaluate our model under the fine-tuning protocol on SSV2, Kinetics-400, UCF-101, and HMDB-51. Besides pre-training on YT-Temporal, we further follow recent works~\cite{mcq,miles} to jointly post-pretrain our model on \textbf{Google Conceptual Captions} (CC3M) and \textbf{WebVid-2M}. Their texts are harvested from the web in the form of a single caption. Since there is no timestamp-annotated text on CC3M and WebVid-2M, we only adopt the contrastive object. As listed in Table~\ref{tab: ssv2_k400} and Table~\ref{tab: ucf_hmdb}, the recognition capability of our model is comparable to previous works as we achieve state-of-the-art or competitive accuracy, while retaining strong transferability. Additionally, the video-text alignment methods show inferior performance on SSV2 since they are devoted to associating the vision patterns with language concepts, without fully exploiting the temporal information. By contrast, our TVTS achieves satisfactory performance via strengthening the learning of spatiotemporal representations.

\vspace{-1em}
\subsubsection{Text-to-Video Retrieval}
\vspace{-0.5em}
As we preserve a global video-transcript contrastive loss to ease the ordering task via learning semantically meaningful video representations, it is natural to ask if the semantic-aware video representations can also benefit retrieval. Hence we conduct text-to-video retrieval under the fine-tuning protocol. As reported in Table~\ref{tab: retri}, our model achieves SOTA performance. The promising results show that our TVTS can also learn the association between video patterns and language semantics.

\vspace{-0.5em}
\section{Conclusion}
\vspace{-0.5em}
In this work, we for the first time leverage script knowledge that is naturally tied to the video to facilitate spatiotemporal representation learning. We introduce a novel pretext task dubbed \emph{Turning to Video for Transcript Sorting} (TVTS), which regularizes the model to learn transferable video representations for spatial and temporal reasoning. Extensive evaluations on downstream video tasks show the great superiority of our method. 

\begin{scriptsize}
\begin{spacing}{1.0}
{\flushleft \bf Acknowledgement.}
This work is supported in part by the National Natural Science Foundation of China under grant 62171248, the PCNL KEY project (PCL2021A07), Guangdong Provincial Key Laboratory of Novel Security Intelligence Technologies (2022B1212010005), Guangdong Basic and Applied Basic Research Foundation under grant 2021A1515110066, the GXWD 20220811172936001, the Shenzhen Science and Technology Innovation Commission (Research Center for Computer Network (Shenzhen) Ministry of Education), and Shenzhen Science and Technology Program under Grant JCYJ20220818101012025.
\end{spacing}
\end{scriptsize}


{\small
\bibliographystyle{ieee_fullname}
\bibliography{egbib}
}

\clearpage
\appendix

\section{Downstream Datasets}
\subsection{Action Recognition}\label{appendix: action_data}
The statistics of our downstream action recognition datasets are listed as follows:
(a) \textbf{Something-Something V2} (SSV2)~\cite{ssv2} is a large-scale dataset that shows humans performing pre-defined basic actions with everyday objects.
It consists of 169K training videos and 20K validation videos belonging to 174 fine-grained action classes.
(b) \textbf{Kinetics-400}~\cite{kinetics} contains 240K training videos and 20K validation videos belonging to 400 classes.
(c) \textbf{UCF-101}~\cite{ucf} contains 9.5K/3.5K training and validation videos with 101 action classes.
(d) \textbf{HMDB-51}~\cite{hmdb} contains 3.5K/1.5K training and validation videos with 51 action classes.

\subsection{Text-to-Video Retrieval}\label{appendix: retri_data}
The statistics of our downstream text-to-video retrieval datasets are listed as follows:
(a) \textbf{MSR-VTT}~\cite{msrvtt} contains 10K YouTube videos with 200K descriptions. 
Following~\cite{frozen}, we train on the training and validation set consisting of 9K videos and evaluate on the 1K-A test set.
(b) \textbf{MSVD}~\cite{msvd} contains 1,970 YouTube videos with 80K descriptions, where each video has around 40 sentences.
We adopt the official split~\cite{frozen}, in which 1200, 100, and 670 videos are used for training, validation, and testing respectively.
(c) \textbf{DiDeMo}~\cite{didemo} contains 10K Flickr videos with 40K sentences.
We follow~\cite{frozen,mcq,miles} to evaluate paragraph-to-video retrieval, \ie, we concatenate all sentences for a video to form a single query.
Specifically, we directly use the whole video without cropping the localized moments (as done by \cite{frozen,mcq,miles}).
(d) \textbf{LSMDC}~\cite{lsmdc} consists of 118,081 video clips harvested from 202 movies.
We adopt the split of \cite{frozen}, where the validation and test set has 7,408 and 1,000 videos respectively.

\section{Implementation Details}\label{appendix: hyper}
As some of the YT-Temporal dataset's video sources, \eg, YouTube, are overlapped with those of downstream datasets, 
we have carefully checked that there is no data leakage between pre-training and downstream datasets by extracting respective frame features with CLIP, calculating their similarity between frame features, and manually examining those with similarity above the threshold.

Our training hyper-parameters are listed in Table~\ref{tab: hyper_pre} and Table~\ref{tab: hyper_ft}. We mostly follow the setting of~\cite{VideoMAE} for convenience. Carefully tuning these parameters may yield better performance.

\begin{table}[t]
\setlength\tabcolsep{8pt}
\centering
\caption{The pre-train and post-pretrain setup.}
\resizebox{1.\columnwidth}{!}{
\begin{tabular}{l|cc}
\toprule
config                                     & pre-train    & post-pretrain  \\
\midrule
optimizer                                  & \multicolumn{2}{c}{AdamW}      \\
learning rate                              & \multicolumn{2}{c}{$1\times10^{-4}$}     \\
batch size                                 & 1024         & 800             \\
training epochs                            & 20           & 12              \\
training frames                            & 16           & 1 $+$ 4             \\
masking ratio                              & 75\%         & 0               \\
input size                                 & \multicolumn{2}{c}{224 $\times$ 224}  \\
patch size, $P$                              & \multicolumn{2}{c}{16}         \\
data augmentation                          & \multicolumn{2}{c}{RandomCrop} \\
hidden state dimension, $D_h$               & \multicolumn{2}{c}{768}        \\
common space dimension, $D$                  & \multicolumn{2}{c}{256}        \\
temperature parameter, $\tau$ & \multicolumn{2}{c}{0.05}       \\
\bottomrule
\end{tabular}}
\label{tab: hyper_pre}
\end{table}

\begin{table}[t]
\setlength\tabcolsep{8pt}
\centering
\caption{The linear probe and fine-tuning setup.}
\resizebox{1.\columnwidth}{!}{
\begin{tabular}{l|cc}
\toprule
config            & linear probe    & fine-tuning               \\
\midrule
optimizer         & SGD             & AdamW                     \\
learning rate     & 0.1             & 0.001                     \\
batch size        & 384             & 384                       \\
training epochs   & 100             & 50 (SSV2), 100 (Others)    \\
training frames   & \multicolumn{2}{c}{16}                      \\
clips $\times$ crops     & \multicolumn{2}{c}{5 $\times$ 3 (K400), 2 $\times$ 3 (Others)} \\
data augmentation & \multicolumn{2}{c}{CenterCrop}              \\
\bottomrule
\end{tabular}}
\label{tab: hyper_ft}
\end{table}

\begin{table*}[t]
\centering
\caption{The full results for text-to-video retrieval on MSR-VTT, DiDeMo, LSMDC, and MSVD.}
\resizebox{.9\linewidth}{!}{
\begin{tabular}{ccccccccccc}
\toprule
\multicolumn{5}{c}{MSR-VTT}            &  & \multicolumn{5}{c}{DiDeMo}             \\
\cmidrule(l){1-5} \cmidrule(l){7-11}
Method     & R@1  & R@5  & R@10 & MedR &  & Method     & R@1  & R@5  & R@10 & MedR \\
\cmidrule(l){1-5} \cmidrule(l){7-11}
NoiseEst~\cite{noiseest}   & 17.4 & 41.6 & 53.6 & 8.0    &  & HERO~\cite{hero}       & 2.1  & -    & 11.4 & -    \\
MMT~\cite{mmt}        & 26.6 & 57.1 & 69.6 & 4.0    &  & CE~\cite{ce}         & 16.1 & 41.1 & 82.7 & 8.3  \\
SupportSet~\cite{supportset} & 30.1 & 58.5 & 69.3 & 3.0    &  & ClipBert~\cite{clipbert}   & 20.4 & 48.0   & 60.8 & 6.0    \\
Frozen~\cite{frozen}     & 31.0   & 59.5 & 70.5 & 3.0    &  & Frozen~\cite{frozen}     & 31.0   & 59.8 & \textbf{72.4} & 3.0    \\
Ours       & \textbf{34.6} & \textbf{61.5} & \textbf{72.2} & \textbf{3.0}    &  & Ours       & \textbf{32.4} & \textbf{59.8} & 71.7 & \textbf{3.0}    \\
\cmidrule(l){1-5} \cmidrule(l){7-11}
\multicolumn{5}{c}{LSMDC}              &  & \multicolumn{5}{c}{MSVD}               \\
\cmidrule(l){1-5} \cmidrule(l){7-11}
Method     & R@1  & R@5  & R@10 & MedR &  & Method     & R@1  & R@5  & R@10 & MedR \\
\cmidrule(l){1-5} \cmidrule(l){7-11}
NoiseEst~\cite{noiseest}   & 6.4  & 19.8 & 28.4 & 39.0   &  & NoiseEst~\cite{noiseest}   & 20.3 & 49.0   & 63.3 & 6.0    \\
MMT~\cite{mmt}        & 12.9 & 29.9 & 40.1 & 19.3 &  & SupportSet~\cite{supportset} & 28.4 & 60.0   & 72.9 & 4.0    \\
Frozen~\cite{frozen}     & 15.0   & 30.8 & 39.8 & 20.0   &  & Frozen~\cite{frozen}     & 45.6 & \textbf{79.8} & \textbf{88.2} & 2.0    \\
Ours       & \textbf{17.2} & \textbf{32.8} & \textbf{41.7} & \textbf{17.0}   &  & Ours       & \textbf{45.9} & 76.7 & 85.4 & \textbf{2.0}   \\
\bottomrule
\end{tabular}}
\label{tab: all_full}
\end{table*}

\begin{table}[t]
\setlength\tabcolsep{4pt}
\centering
\caption{Experiments on SVO Probes, a recently proposed benchmark for the subject, verb, and object understanding in static images. Our pre-trained model can better reason about the dynamic context behind the given images. We do not compare with SOTA spatiotemporal representation learning methods, \eg, VideoMAE, since they cannot perform text-to-video retrieval.}
\resizebox{1.\columnwidth}{!}{
\begin{tabular}{c|ccc|ccc|ccc}
\toprule
$\rho$       & \multicolumn{3}{c}{0.2} & \multicolumn{3}{c}{0.25} & \multicolumn{3}{c}{0.3} \\
\midrule
Method & subj   & obj    & verb  & subj   & obj    & verb   & subj   & obj    & verb  \\
\midrule
Frozen & 0.56   & 0.61   & 0.54  & 0.58   & 0.66   & 0.56   & 0.62   & 0.72   & 0.58  \\
Ours   & \textbf{0.59}   & \textbf{0.65}   & \textbf{0.59}  & \textbf{0.64}   & \textbf{0.70}   & \textbf{0.62}   & \textbf{0.68}   & \textbf{0.76}   & \textbf{0.63}  \\
\bottomrule
\end{tabular}}
\label{tab: svo}
\end{table}

\begin{table}[t]
\centering
\caption{The top-1 accuracy \wrt different contrastive formulation on SSV2 under the fine-tuning protocol.}
\resizebox{.85\columnwidth}{!}{
\begin{tabular}{cccccc}
\toprule
Name & Formulation              & $\mathcal{L}_{\text{base}}$ & $\mathcal{L}_{\text{sort}}$ & SSV2 & Gain \\
\midrule
M$_1$ & \multirow{2}{*}{MERLOT} & \cmark       & \xmark & 66.2 & \multirow{2}{*}{+0.9} \\
M$_2$ &                         & \cmark       & \cmark       & 67.1 \\
\midrule
M$_3$ & \multirow{2}{*}{Ours}   & \cmark       & \xmark       & 67.0 & \multirow{2}{*}{+1.5} \\
M$_4$ &                         & \cmark       & \cmark       & \textbf{68.5} \\
\bottomrule
\end{tabular}}
\label{tab:con_form}
\end{table}

\begin{table}[t]
\centering
\caption{The sort accuracy \wrt different sort modules. T (V) denotes the transcript (video) representation; RG refers to random guessing, and SortTSF refers to the sort transformer.}
\resizebox{.85\columnwidth}{!}{
\begin{tabular}{cccc}
\toprule
Sort Source & Transcripts, $K$      & Sort Module & Accuracy \\
\midrule
T           & \multirow{3}{*}{4} & RG          & 0.4\%   \\
T           &                    & SortTSF     & 0.5\%   \\
T + V         &                    & SortTSF     & \textbf{21.5\%} \\
\bottomrule
\end{tabular}}
\label{tab:sort_acc}
\end{table}

\section{Additional Experiments}\label{appendix: add_exp}
\subsection{Full Results for Text-to-Video Retrieval}
We compare our method with seven state-of-the-art methods~\cite{noiseest,mmt,supportset,frozen,hero,ce,clipbert}. The full Recall@K and MedR results are reported in Table~\ref{tab: all_full}.
Our model achieves state-of-the-art or competitive performance on all datasets. It shows that our TVTS is capable of learning the association between video patterns and language semantics.

\subsection{SVO-Probes Test}

Our model can also be well transferred to understand static images and reason about the dynamic context behind them.
To evaluate such an ability, we conduct experiments on the recently proposed SVO Probes~\cite{svo}, a zero-shot test benchmark for \textit{subject}, \textit{verb}, and \textit{object} understanding in the image field.
In SVO Probes, each sentence is tied with a positive and a negative image, in which the positive image has consistent semantics, \ie subject, verb, and object, with the sentence, while the negative image substitutes one of the three concepts but keeps the remaining two unchanged. The objective is to test whether a model can correctly identify the positive image given a query sentence. 
We treat it as a text-image retrieval task,
\ie given the text and image embedding, if their cosine similarity surpasses a certain threshold $\rho$, we consider the image positive. We report the precision results in terms of different values of $\rho$, shown in Table~\ref{tab: svo}. Our model reaches higher precision on all concepts, which implies our learned spatiotemporal representations have strong out-of-the-box capabilities.

\subsection{Ablation Study (Cont.)}

\begin{figure}[t]
	\centering
	\includegraphics[width=1.\columnwidth]{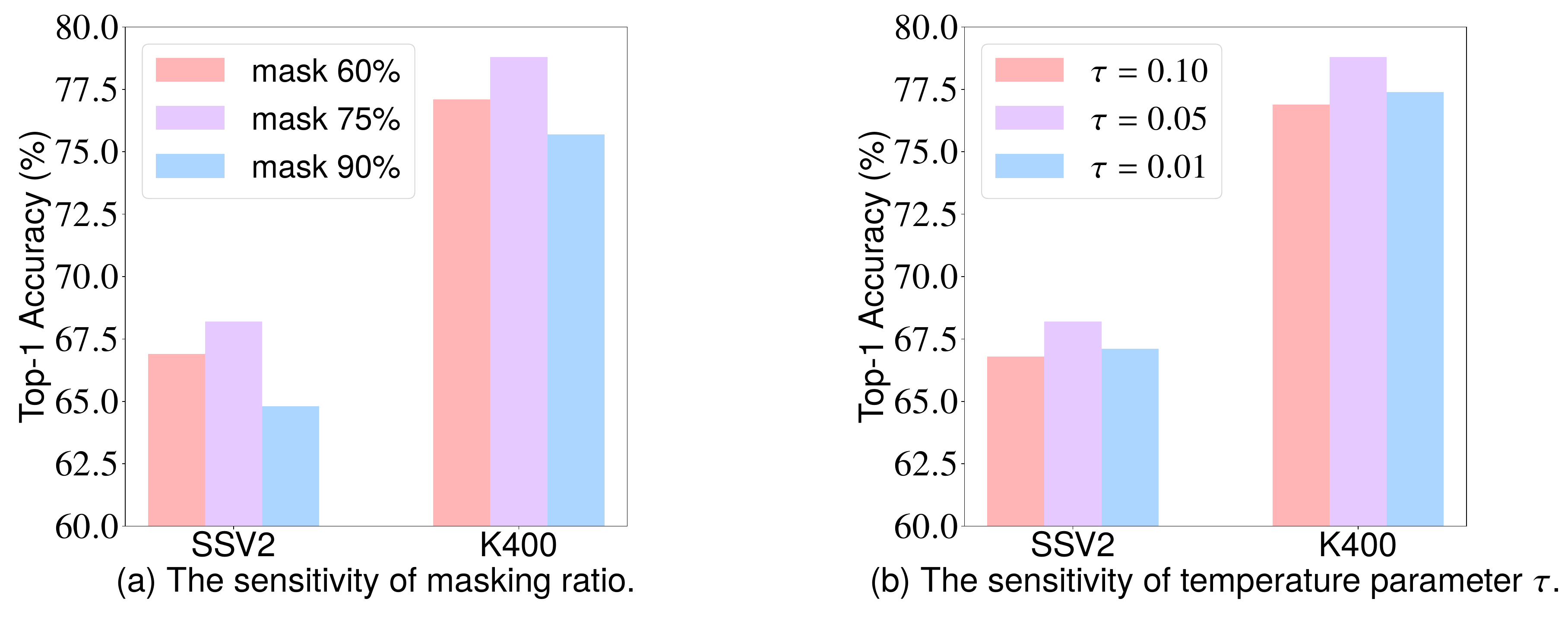}
	\caption{(a) The top-1 accuracy \wrt different masking ratio.
	(b) The top-1 accuracy \wrt different temperature parameter $\tau$.}
	\label{fig: params}
\end{figure}

\begin{figure*}[t]
	\centering
	\includegraphics[width=1.\linewidth]{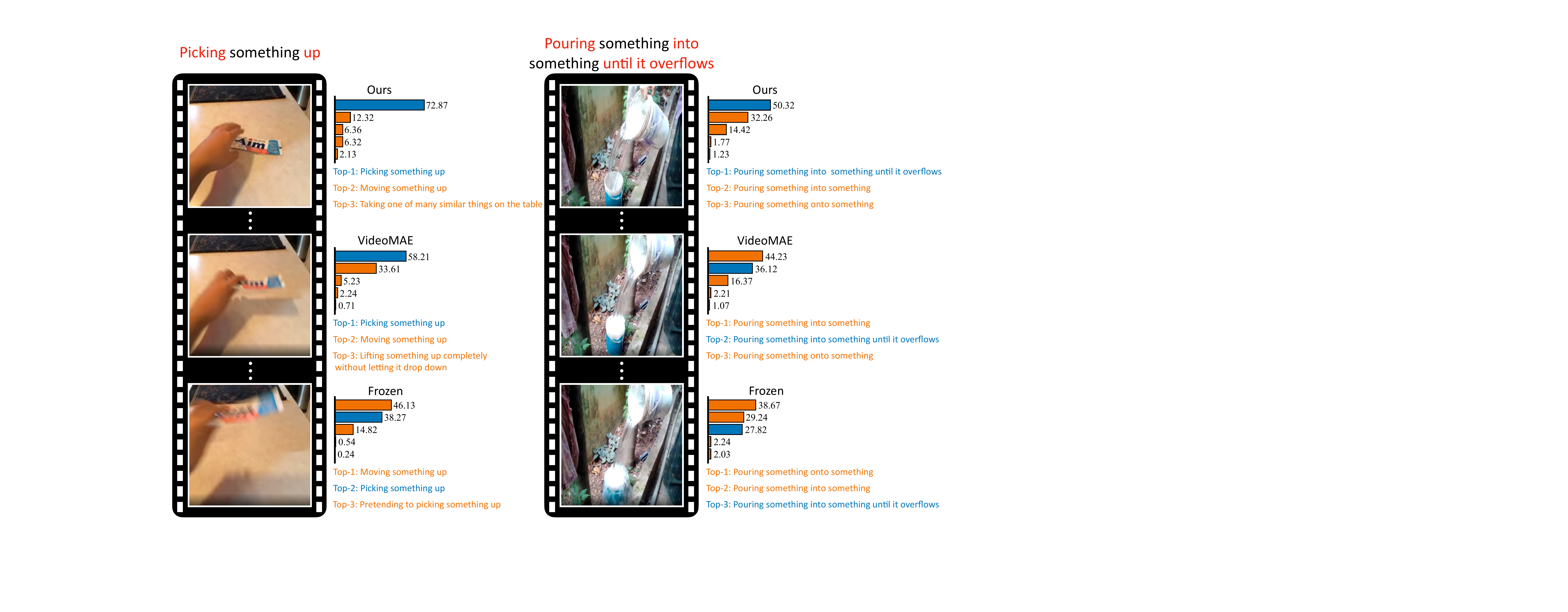}
	\caption{Visualization of the top-5 prediction scores on SSV2, we normalize the scores to make their summation 100\%. The blue and orange rows denote the scores of the right and wrong classes, respectively.}
	\label{fig: top-5}
\end{figure*}

{\flushleft \bf Contrastive Formulation.}
Since MERLOT~\cite{merlot} formulates the contrastive objective by frame-transcript matching, we further investigate how much this change in the proposed approach from MERLOT contributes to the improved performance.
Specifically, we replace the contrastive formulation of $\mathcal{L}_{\text{base}}$ with that of MERLOT, and the results are reported in Table~\ref{tab:con_form}.
The accuracy slightly degrades due to mismatches between single frames and noisy transcripts, but the sorting task still boosts video representations, given the gains when plugging $\mathcal{L}_{\text{sort}}$.

{\flushleft \bf Sort Accuracy.}
To prevent the model from learning shortcuts, \ie, memorizing orders from text alone, we stop the gradients of sorting loss from flowing toward encoding transcript features.
To verify it, we test the accuracy of transcript sorting using our pre-trained model in Table~\ref{tab:sort_acc}, where the expectation of random guessing accuracy is 0.4\% (1/$4^4$).
Sorting the text alone almost fails, while sorting text via resorting to video features achieves 21.5\% accuracy.
It implies the sorting task is solved by promoting video understanding instead of learning shortcuts.

{\flushleft \bf Masking Ratio.}
We compare different masking ratios for TVTS in Figure~\ref{fig: params}(a). Both lower (60\%) and higher (90\%) masking ratio drop performance than our method with 75\% ratio, because a lower masking ratio brings in temporal redundancy, while a higher ratio leads to the extremely limited knowledge to perform TVTS.


{\flushleft \bf Temperature Parameter.}
We also investigate the influence of the temperature parameter $\tau$ in $\mathcal{L}_{\text{base}}$ in Figure~\ref{fig: params}(b). A smaller $\tau$ makes the model focus more on the hard negative samples, but it also increases the difficulty of convergence. We set $\tau=0.05$ for its best performance.

{\flushleft \bf Visualization.}
To demonstrate the superiority of our learned spatiotemporal representation intuitively, we randomly pick two videos in SSV2 and illustrate the top-5 prediction scores \wrt our method, VideoMAE and Frozen in Figure~\ref{fig: top-5}. 
Our method predicts the highest score for the right class. In the first column, we need to distinguish the action ``picking'' from other similar actions such as ``moving'', which requires fine-grained temporal reasoning ability. In the second column, the model must extract both the spatial and temporal information to classify the video as the category containing ``into'' and ``until it overflows''. Only our method classifies the video correctly, while VideoMAE and Frozen make mistakes due to a lack of spatiotemporal modeling ability.

\end{document}